\documentclass{article}

\usepackage[accepted]{icml2020}

\usepackage[utf8]{inputenc}         
\usepackage[T1]{fontenc}            
\usepackage{hyperref}               
\usepackage{url}                    
\usepackage{booktabs}               
\usepackage{amsfonts}               
\usepackage{amssymb}                
\usepackage{mathtools}              
\usepackage{nicefrac}               
\usepackage{microtype}              
\usepackage{xcolor}                 
\usepackage{import}                 
\usepackage{amsmath}                
\usepackage{graphicx}               
\usepackage{tabularx}               
\usepackage{pgf, tikz}              
\usepackage[shortlabels]{enumitem}  
\usepackage{natbib}                 
\usepackage{multicol}               
\usepackage{wrapfig}                
\usepackage{subcaption}             
\usepackage{fancyhdr}               
\usepackage{lastpage}               
\usepackage{epigraph}               
\usepackage{scrextend}              
\usepackage{multicol}               
\usepackage{wrapfig}                
\usepackage{subcaption}             
\usepackage{scrextend}              
\usepackage{multirow}               
\usepackage{booktabs}

\usepackage{custom_configs}         

\icmltitlerunning{Super-resolution Variational Auto-Encoders}

\begin{document}
    \twocolumn[
    \icmltitle{Super-resolution Variational Auto-Encoders}
    \icmlsetsymbol{equal}{*}
    \begin{icmlauthorlist}
    \icmlauthor{Ioannis Gatopoulos}{uva,vu,bc}
    \icmlauthor{Maarten Stol}{bc}
    \icmlauthor{Jakub M. Tomczak}{vu}
    \end{icmlauthorlist}
    \icmlaffiliation{uva}{Department of Computer Science, University of Amsterdam, Amsterdam, The Netherlands}
    \icmlaffiliation{vu}{Department of Computer Science, Vrije Universiteit Amsterdam, Amsterdam, The Netherlands}
    \icmlaffiliation{bc}{BrainCreators B.V., Amsterdam, The Netherlands}
    \icmlcorrespondingauthor{Ioannis Gatopoulos}{ioannis.gatopoulos@gmail.com}
    \icmlkeywords{Machine Learning, Generative Modeling, ICML, INNF}
    \vskip 0.3in
    ]
    \printAffiliationsAndNotice{}
    \begin{abstract}
    The framework of variational autoencoders (VAEs) provides a principled method for jointly learning latent-variable models and corresponding inference models.
However, the main drawback of this approach is the blurriness of the generated images.
Some studies link this effect to the objective function, namely, the (negative) log-likelihood (\textit{nll}).
Here, we propose to enhance VAEs by adding a random variable that is a downscaled version of the original image and still use the log-likelihood function as the learning objective.
Further, by providing the downscaled image as an input to the decoder, it can be used in a manner similar to the super-resolution.
We present empirically that the proposed approach performs comparably to VAEs in terms of the nll, but it obtains a better Fréchet Inception Distance (\textit{FID}) score in data synthesis.

    \end{abstract}
    
    \section{Introduction}
    \label{sec:introduction}
\begin{figure}[t]
    \centering
    \scalebox{1.}{
    \includegraphics[width=0.98\linewidth]{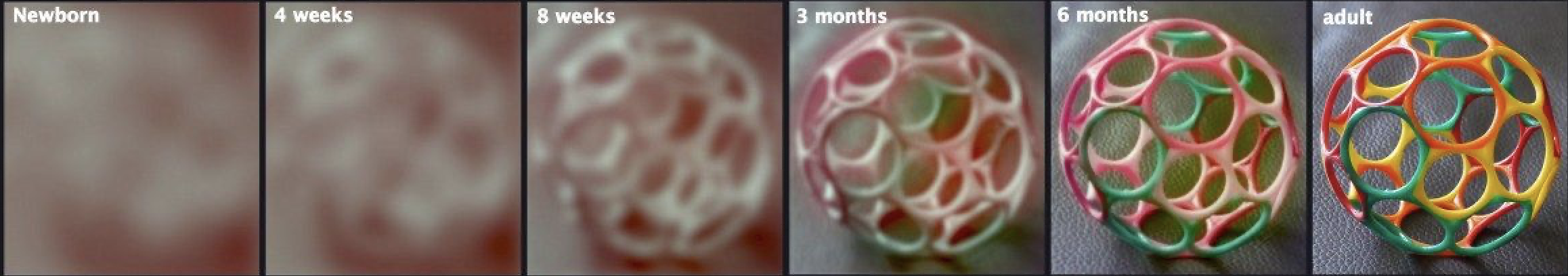}}
    \caption{Infant vision development during the first six months. The previous processed information is enriched with new signals in order for the sensory systems to be able to analyze high fidelity signals \citep{infant_vision}.}
    \label{fig:human_vision}
\end{figure}

    Unlike many other sensory systems, the human visual system (i.e. the components from the eye to neural circuits) develops largely after birth, especially in the first few years of life \citep{infant}. In the beginning, even though the visual structures are fully present, they are still immature in their potentials. As the neural circuits adapt to natural light, they learn to enhance the already known signals with the new information that is becoming available (Figure \ref{fig:human_vision}).
Furthermore, \cite{obj_rec} suggested that the human visual system for object recognition tasks initially starts with the skeletal structure of the object and then maps other properties, such as textures and colors, onto it before it is classified. 
It seems that humans reinforce their capabilities by sequentially applying new content of information over time and some specific processes, like object detection, are divided into two or more simpler tasks.

Inspired by this learning procedure, we formulate a generative model that mimics, to some extent, the human visual process. 
Specifically, we enhance the framework of Variational Auto-Encoder (VAEs) by introducing a downscaled representation of the image as a random variable, and utilize it in a super-resolution manner \citep{chang2004super, dong2015image, freeman2002example} to generate high quality images.
As a result, we obtain a two-level VAE with three latent variables, where one is the downscaled version of the original image.

In summary, our contributions are as follows:
\begin{itemize}[label=\raisebox{0.45ex}{\tiny$\bullet$}]
    \item We present a powerful Variational Auto-Encoder that consists of a novel DenseNet-based encoder, a DenseNet-based decoder, and a flow-based prior. It achieves SOTA in terms of the log-likelihood function among singe-leveled VAEs.
    \item We propose a new class of VAEs that contain a super-resolution part for generating crisp images, and is still trained using the log-likelihood objective.
    \item We present empirical results on CIFAR-10 and ImageNet32 where our approach achieves descent scores in terms of the bits per dimension (bpd) on CIFAR-10 and ImageNet32, and impressive FID scores.
\end{itemize}

    \section{Variational Auto-Encoders}
    \label{sec:background.tex}

Let $\mathbf{X} = \{ \mathbf{x}_1, ..., \mathbf{x}_N\}$ with $\mathbf{x}_n \in \mathbb{R}^{\mathrm{D}}$ be the observable data that we wish to model. 
Further, we consider a latent variable model with latent (unobserved) variables $\mathbf{z} \in \mathbb{R}^{\mathrm{M}}$, $p_{\vartheta}(\mathbf{x}) = \int p_{\vartheta}(\mathbf{x}, \mathbf{z}) \mathrm{d} \mathbf{z}$, where $\vartheta$ denotes parameters.
We consider the optimization through maximum likelihood estimation (MLE) of $p_{\vartheta}(\mathbf{x})$, however, it becomes infeasible due to the intractability of the integration at hand. 
One possible way of overcoming this issue and obtaining a highly scalable framework is by introducing an amortized variational family $\mathcal{Q}$ in order to identify its member $q_{\phi}(\mathbf{z}|\mathbf{x})$ that minimizes the Kullback-Leibler divergence to the \textit{real} posterior $p(\mathbf{z}|\mathbf{x})$. 
In consequence, we derive a tractable objective function, namely the \textit{evidence lower bound} (ELBO) \citep{Jordan1998}:
\begin{align}
    \log &p_{\vartheta}(\mathbf{x})
    \geq
    \mathbb{E}_{q_{\phi}(\mathbf{z}|\mathbf{x})}
    \log \left[
    \frac{p_{\theta}(\mathbf{x}, \mathbf{z})}
    {q_{\phi}(\mathbf{z}|\mathbf{x})} \right]
    \nonumber
    \\&=
    \mathbb{E}_{q_{\phi}(\mathbf{z} | \mathbf{x})}
    \left[\log p_{\theta}(\mathbf{x} | \mathbf{z})
    - \log q_{\phi}(\mathbf{z} | \mathbf{x}) +
    \log p_{\lambda}(\mathbf{z})
    \right]
    \nonumber
    \\&\equiv
    \mathcal{L}(\theta, \phi, \lambda),
\end{align}
where $q_{\phi}(\mathbf{z} | \mathbf{x})$ is the variational posterior (or the \textit{encoder}), $p_{\theta}(\mathbf{x} | \mathbf{z})$ is the likelihood function (or the \textit{decoder}) and $p_{\lambda}(\mathbf{z})$ is the \textit{prior} over the latent variables, parameterized by and $\phi, \theta$ and $\lambda$ respectively. 
The optimization is done efficiently by computing the expectation by Monte Carlo integration while exploiting the \textit{reparameterization trick} in order to obtain an unbiased estimator of the gradients. 
This generative model framework is known as \textit{Variational Auto-Encoder} (VAE) \citep{kingma2013autoencoding, rezende2014stochastic}.


\paragraph{VAE with a bijective prior} Even though the lower-bound suggests that the prior plays a crucial role in improving the variational bounds, usually it is modelled by a fixed distribution (i.e., a standard multivariate Gaussian). 
While being relatively simple and computationally cheap, a fixed prior is known to result in over-regularized models that tend to ignore more of the latent dimensions \citep{burda2015importance, tomczak2017vae}. 
Moreover, as the objective function is optimised to match the variational posterior with the prior, \cite{rosca2018distribution} argued that even if the former becomes the optimal one, namely the \textit{aggregated posterior}, it may still not match a unit Gaussian distribution. 

However, it is possible to obtain a rich, multi-modal prior distribution $p(\mathbf{z})$ by using a \textit{bijective model}. 
Formally, given a latent code $\mathbf{z} \sim q_Z(\mathbf{z}| \mathbf{x})$, a base distribution $p_V(\mathbf{v})$ on a latent variable $\mathbf{v} \in V$, and $f: V \xrightarrow{} Z$ consisting of a sequence of $L$ diffeomorphic transformations\footnote{That is, invertible and differentiable transformations.}, where $f_i(\mathbf{v}_{i-1}) = \mathbf{v}_{i}$, $\mathbf{v}_{0} = \mathbf{v}$ and $\mathbf{v}_{L} = \mathbf{z}$, the sequential use of the \textit{change of variable} can be used to express the distribution of $\mathbf{z}$ as a function of $\mathbf{v}$ as follows:
\begin{align}
    \log p_{Z}(\mathbf{z})
    =
    \log p_{V}(\mathbf{v}) - \sum_{i=1}^{L} \log \left| \frac{\partial f_i(\mathbf{v}_{i-1})}{\partial \mathbf{v}_{i-1}} \right| ,
\end{align}
where $ \left|\frac{\partial f_i(\mathbf{v}_{i-1})}{\partial \mathbf{v}_{i-1}} \right|$ is the Jacobian-determinant of the $i^{th}$ transformation.

Thus, using the transformed prior we end up with the following training objective function:
\begin{align}
    \mathcal{L}\left(\theta, \phi, \lambda\right)
    =& \,
    \mathbb{E}_{q_{\phi}(\mathbf{z} | \mathbf{x})}
    \Big[
    \log p_{\theta}(\mathbf{x} | \mathbf{z})
    - 
    \log q_{\phi}(\mathbf{z} | \mathbf{x}) \,+
    \nonumber
    \\+&
    \log p_{V}(\mathbf{v}_{0})
    \, + \,
    \sum_{i=1}^{L}
    \log\left|\frac{\partial f^{-1}_i(\mathbf{v}_{i})}{\partial \mathbf{v}_{i}} \right| \Big] .
\end{align}

In this paper, we utilize RealNVP \citep{dinh2016density} as the prior, however, any other flow-based model could be used \citep{berg2018sylvester, kingma2018glow}.

    \section{Our method}
    \label{sec:method.tex}

\subsection{Model formulation}
\begin{figure}[t]
\centering
\scalebox{0.75}{
\begin{subfigure}{.5\textwidth}
    \includegraphics[width=1\linewidth]{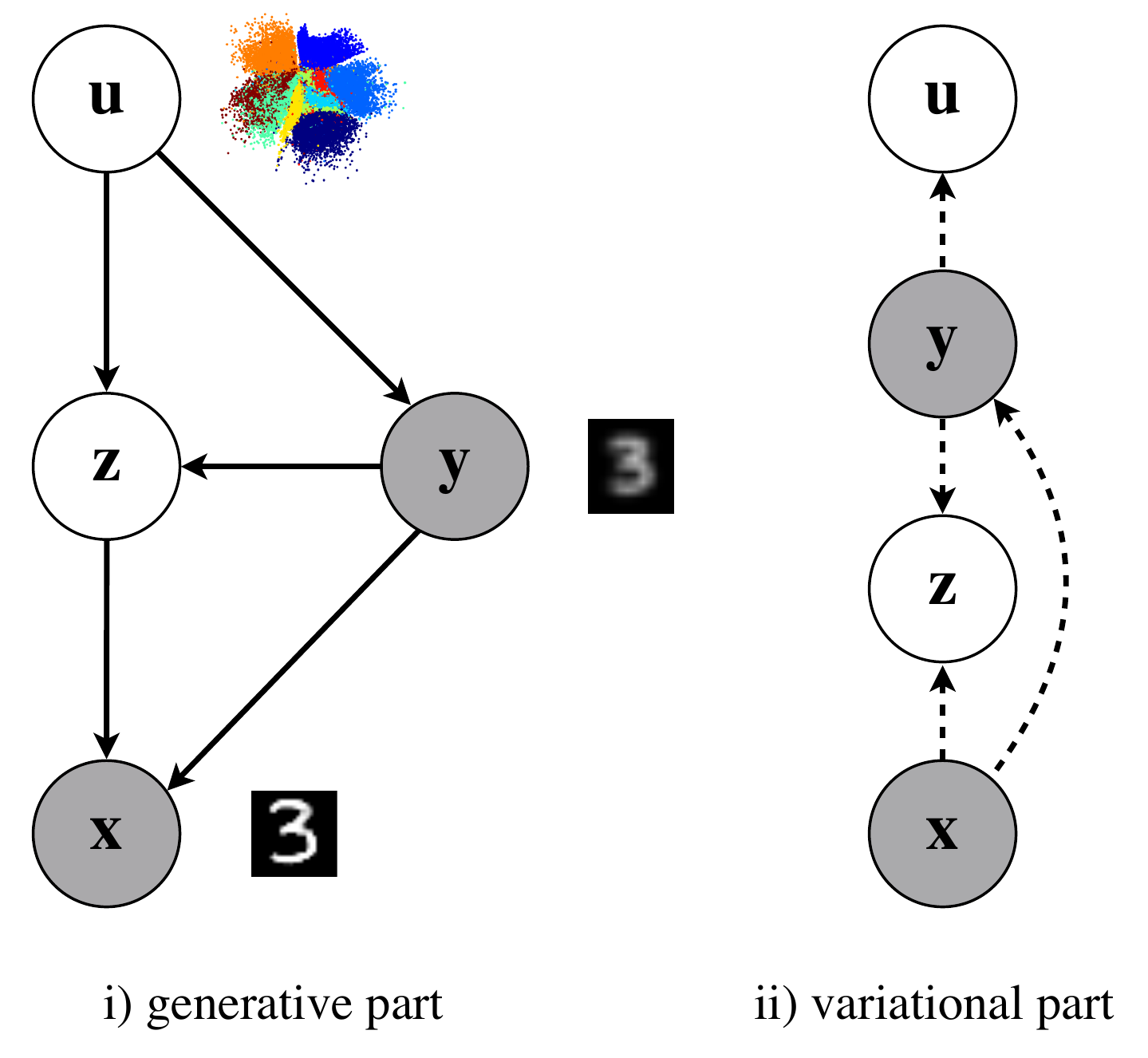}
\end{subfigure}}
\caption[Bayesian graph of proposed models]{Stochastic dependencies of the proposed model. Our approach takes advantage of a compressed representation $\mathbf{y}$ of the data in the variational part, that is then utilized in the super-resolution in the generative part.}
\label{fig:bg_model}
\end{figure}

Let us introduce an additional variable $\mathbf{y}\in \mathbb{R}^{\mathrm{C}}$ that is a compressed representation of $\mathbf{x}$\footnote{Here, we consider a downscaled $\mathbf{x}$ however, our framework allows for any compressed transformation of the original data.}, where $C\leq D$. Further, let $\mathbf{u} \in \mathbb{R}^{K}$ and $\mathbf{z} \in \mathbb{R}^{M}$ be two stochastic latent variables that interact with the above observed ones in a way that is presented in Figure \ref{fig:bg_model}. 


From the dependencies of the considered probabilistic graphical model, we can write the joint probability as follows:
\begin{align*}
    p(\mathbf{x}, \mathbf{y}, \mathbf{z}, \mathbf{u})
    =
    p(\mathbf{x}|\mathbf{y}, \mathbf{z}) \,
    p(\mathbf{z}|\mathbf{y}, \mathbf{u}) \,
    p(\mathbf{y}|\mathbf{u}) \,
    p(\mathbf{u}).
\end{align*}
Then, we define the amortized variational posterior of $p(\mathbf{y}, \mathbf{z}, \mathbf{u}| \mathbf{x})$ as follows:
\begin{align*}
     q(\mathbf{y}, \mathbf{z}, \mathbf{u}| \mathbf{x}) = q(\mathbf{z}|\mathbf{y}, \mathbf{x}) \, q(\mathbf{u}|\mathbf{y}) \, q(\mathbf{y}|\mathbf{x}) \equiv q(\mathbf{w}| \mathbf{x})
\end{align*}
where $\mathbf{w} = \{\mathbf{y}, \mathbf{z}, \mathbf{u}\}$, and derive the corresponding lower bound of the likelihood function in the following manner:
\begin{align*}
    \log p(\mathbf{x})
    &\ge
    \E\mathord{_{q(\mathbf{w})}} \log \frac{p(\mathbf{x}, \mathbf{w})}{q(\mathbf{w})} 
    \\ &= 
    \E\mathord{_{q(\mathbf{w})}}
    \Big[
    \log p(\mathbf{x}, \mathbf{w})
    \Big]
    -
    \E\mathord{_{q(\mathbf{w})}}
    \Big[
    \log q(\mathbf{w})
    \Big]
    \numberthis \label{eq:elbo1}
    \\&\equiv
    \mathcal{L}(\mathbf{x}).
\end{align*}
After expanding and rearranging the above objective function (please see \ref{sec:appendix_A1} for full derivation), we obtain:
\begin{align}\label{eq:our_elbo}
    \mathcal{L}(\mathbf{x})
    =&
    \mathbb{E}_{q(\mathbf{z}| \mathbf{x}, \mathbf{y}) \, q(\mathbf{y}|\mathbf{x})} \log 
    p_{\theta}(\mathbf{x}| \mathbf{y}, \mathbf{z}) +
    \nonumber
    \\&
    - \E\mathord{_{q(\mathbf{u}|\mathbf{y})}} \KL({q(\mathbf{y}| \mathbf{x}})||p_{\theta}(\mathbf{y}| \mathbf{u}) ) +
    \nonumber
    \\&  
    -  \mathbb{E}_{q(\mathbf{u}|\mathbf{y}) q(\mathbf{y}|\mathbf{x})}\KL({q(\mathbf{z}| \mathbf{x}, \mathbf{y}) || p(\mathbf{z}| \mathbf{y}, \mathbf{u}})) +
    \nonumber
    \\&
    - \mathbb{E}_{q(\mathbf{y}|\mathbf{x})}\KL({q(\mathbf{u}| \mathbf{y}})||p(\mathbf{u})) ,
\end{align}
where $\KL(\cdot||\cdot)$ denotes the Kullback-Leibler divergence.

\subsection{Properties}

There are two main properties that we are going to take advantage of.


\subsubsection*{A. \, If $q(\mathbf{y}|\mathbf{x})$ is both \textbf{deterministic} and \textbf{discrete}, then $\E\mathord{_{q(\mathbf{y}|\mathbf{x})}} \Big[ \log q(\mathbf{y}|\mathbf{x}) \Big] =0$.}

Dependence between two random variables can take a variety of forms, of which \textit{stochastic independence} and \textit{functional dependence} can be argued to be most opposite in character.
In the former case, neither of the variables provide any information about each other, whereas in the latter, there is a full determination.
Even though the proposed framework allows to model $q(\mathbf{y}|\mathbf{x})$ as a stochastic dependency, the choice of a deterministic relationship is more attractive, as the transformations to a compressed representation are usually available (e.g., a downscaled image), the optimization process would be faster, easier and the model overall will require less trainable parameters.

We will define this deterministic transformation as a \textit{degenerate} probability distribution which provides a way to deal with constant values in a probabilistic framework. It trivially gives rise to a probability mass function satisfying $P(\Omega)=1$ and has an expectation of a constant value $c \in \mathbb{R}$, a variance of $0$ and most importantly, its entropy is also equal to $0$. Thus, modelling the distribution $q(\mathbf{y}|\mathbf{x})$ as a discrete degenerate distribution yields:
\begin{align*}
    \mathbb{H}\big[q(\mathbf{y}|\mathbf{x})\big] = 0 
    \Leftrightarrow \mathbb{E}_{q(\mathbf{y}|\mathbf{x})} \big[ - \log q(\mathbf{y}|\mathbf{x}) \big] = 0
\end{align*}
which simplifies the  derived objective function in (\ref{eq:our_elbo}).

\begin{table*}[t]
\caption{Negative log-likelihood for CIFAR-10 and ImageNet32 test set. The dataset marked with the symbol $^{\dag}$ denotes a model trained only on CIFAR-10 and evaluated on ImageNet32. For FID, we provide values obtained on the test set and the training set (in brackets).}
\centering
\scalebox{1.0}{
\begin{tabular}{llcccccc}
\multirow{2}{*}{\textbf{Dataset}} & \multirow{2}{*}{\textbf{Model}} & \multirow{2}{*}{\textit{\begin{tabular}[c]{@{}c@{}}nll\\ (bits/dim)\end{tabular}}} & \multicolumn{2}{c}{\textit{reconstruction loss}} & \multicolumn{2}{c}{\textit{regularization loss}} & \multirow{2}{*}{FID} \\
 &  &  & $\text{RE}_{x}$ & $\text{RE}_{y}$ & $\text{KL}_{z}$ & $\text{KL}_{u}$ &  \\ \hline
\multirow{2}{*}{Cifar10} & VAE & \textbf{3.51} & 5540 & - & 1966 & - & 41.36 (37.25) \\
 & srVAE & 3.65 & 5107 & 1241 & 619 & 819 & \textbf{34.71} (\textbf{29.95}) \\ \hline
\multirow{2}{*}{$\text{ImageNet32}^{\dag}$} & VAE & \textbf{3.80} & 6386 & - & 1707 & - & 51.82 (N/A) \\
 & srVAE & 4.00 & 5907 & 1257 & 597 & 805 & \textbf{45.37} (N/A) \\
 \bottomrule
\end{tabular}}
\label{tab:table_results}
\end{table*}


\subsubsection*{B. \, \textbf{The distribution $q(\mathbf{z}|\mathbf{y}, \mathbf{x})$ can be simplified to $q(\mathbf{z}|\mathbf{x})$.}}

One of the core motivations behind the architecture of the two staged approach is that the latent variable $\mathbf{z}$ will be able to capture the missing information between $\mathbf{x}$ and $\mathbf{y}$. 
While $\mathbf{u}$ would allow to produce the global structure of the data (e.g., a shape of a horse), the variation of $\mathbf{z}$ will alter high-level features (e.g., varying $\mathbf{z}$ will result into a different color of a horse). 
Thus, since $\mathbf{y}$ is a compressed representation of $\mathbf{x}$, it does not introduce any additional information about $\mathbf{z}$ that is not already in $\mathbf{x}$. 
This intuitively allows to model $\mathbf{z}$ only using $\mathbf{x}$, and, in essence, to replace $q(\mathbf{z}|\mathbf{y}, \mathbf{x})$ with $q(\mathbf{z}|\mathbf{x})$.


\subsubsection*{\textbf{Final ELBO}}

With these two properties in mind, the final lower bound of the marginal likelihood of $\mathbf{x}$ is the following:
\begin{align}\label{eq:final_elbo}
    \mathcal{L}(\mathbf{x})
    =&
    \underbrace{\mathbb{E}_{q(\mathbf{z}| \mathbf{x}) \, q(\mathbf{y}|\mathbf{x})} \log 
    p_{\theta}(\mathbf{x}| \mathbf{y}, \mathbf{z})}_{\mathrm{RE}_{x}} +
    \nonumber
    \\&
    + \underbrace{\mathbb{E}_{q(\mathbf{u}|\mathbf{y}) q(\mathbf{y}|\mathbf{x})} \log p_{\theta}(\mathbf{y}| \mathbf{u})}_{\mathrm{RE}_{y}} +
    \nonumber
    \\&  
    - \underbrace{\mathbb{E}_{q(\mathbf{u}|\mathbf{y}) q(\mathbf{y}|\mathbf{x})} \KL({q(\mathbf{z}| \mathbf{x}) || p(\mathbf{z}| \mathbf{y}, \mathbf{u}}))}_{\mathrm{KL}_{z}} +
    \nonumber
    \\&
    - \underbrace{\mathbb{E}_{q(\mathbf{y}|\mathbf{x})}\KL({q(\mathbf{u}| \mathbf{y}})||p(\mathbf{u}))}_{\mathrm{KL}_{u}} .
\end{align}
\subsection{Super-resolution VAE (srVAE)}

We choose the following distributions in our model:

\begin{align*} 
    q_{\phi_1}\left(\mathbf{u} | \mathbf{y}\right.) &=\mathcal{N}\left(\mathbf{u} | \boldsymbol{\mu}_{\phi_1}(\mathbf{y}), \mathrm{diag}\left(\boldsymbol{\sigma}_{\phi_1}(\mathbf{y})\right)  \right.)
    \\
    q(\mathbf{y}|\mathbf{x}) 
    &= \delta(\mathbf{y} = d(\mathbf{x}))
    \\ 
    q_{\phi_2}\left(\mathbf{z} | \mathbf{x}\right.) &=\mathcal{N}\left(\mathbf{z} | \boldsymbol{\mu}_{\phi_2}(\mathbf{x}), \mathrm{diag}\left(\boldsymbol{\sigma}_{\phi_2}(\mathbf{x})\right)  \right.)
    \\
    p_{\lambda}\left(\mathbf{u}\right.) 
    &= p(\mathbf{v}) \, \prod_{i=1}^{L} \Big|\operatorname{det} \frac{\partial f_i(\mathbf{v}_{i-1})}{\partial \mathbf{v}_{i-1}} \Big| ^{-1}
    \\
    p(\mathbf{v})
    &= \mathcal{N}\left(\mathbf{v} | \mathbf{0}, \mathbf{1})\right.
    \\
    p_{\theta_1}\left(\mathbf{y} | \mathbf{u}\right.) 
    &=
    \sum_{i=1}^{K} \pi_{i}^{(\mathbf{u})} \mathrm{Dlogistic}\Big(\mu_{i}^{(\mathbf{u})}, s_{i}^{(\mathbf{u})}\Big)
    \\
    p_{\theta_2}\left(\mathbf{z} | \mathbf{y}, \mathbf{u}\right.) 
    &=
    \mathcal{N}\left(\mathbf{z} | \boldsymbol{\mu}_{\theta_2}(\mathbf{y}, \mathbf{u}), \operatorname{diag}\left(\boldsymbol{\sigma}_{\theta_2}(\mathbf{y}, \mathbf{u})\right.) \right.)
    \\ 
    p_{\theta_3}\left(\mathbf{x} | \mathbf{z}, \mathbf{y}\right.)
    &= 
    \sum_{i=1}^{K} \pi_{i}^{(\mathbf{z}, \mathbf{y})} \mathrm{Dlogistic}\Big(\mu_{i}^{(\mathbf{z}, \mathbf{y})}, s_{i}^{(\mathbf{z}, \mathbf{y})} \Big).
\end{align*}

where $\mathrm{Dlogistic}$ is defined as the discretized logistic distribution \citep{salimans2017pixelcnn}, $\delta(\cdot)$ is the Dirac's delta, and $d(\mathbf{x})$ denotes the downscaling transformation that returns a discrete values. 

In VAEs, it is possible to use the following functionality:
\begin{itemize}[label=\raisebox{0.45ex}{\tiny$\bullet$}]
\item \textbf{Generation:} The model is able to generate new images through the following process: $\mathbf{z} \sim p(\mathbf{z}) \rightarrow \mathbf{x} \sim p(\mathbf{x}|\mathbf{z})$.
\item \textbf{Reconstruction:} The model allows to reconstruct $\mathbf{x}$ by using the following scheme: $\mathbf{x} \rightarrow \mathbf{z} \sim q(\mathbf{z}|\mathbf{x}) \rightarrow \mathbf{x} \sim p(\mathbf{x}|\mathbf{z})$.
\end{itemize}

Interestingly, our approach allows four operations:
\begin{itemize}[label=\raisebox{0.45ex}{\tiny$\bullet$}]
    \item \textbf{Generation:} The model allows to generate novel content by applying the following hierarchical sampling process: 
    $\mathbf{u} \sim p(\mathbf{u}) \xrightarrow{}
    \mathbf{y} \sim p(\mathbf{y}| \mathbf{u})
    \xrightarrow{}
    \mathbf{z} \sim p(\mathbf{z}| \mathbf{u}, \mathbf{y})
    \xrightarrow{}
    \mathbf{x} \sim p(\mathbf{x}| \mathbf{z}, \mathbf{y})$.
    \item \textbf{Conditional Generation} (or \textit{Super-Resolution Generation})\textbf{:} Given $\mathbf{y}$, we can sample the latent codes:
    $ 
    \mathbf{u} \sim q(\mathbf{u}| \mathbf{y})
    \xrightarrow{}
    \mathbf{z} \sim p(\mathbf{z}| \mathbf{y}, \mathbf{u}),
    \xrightarrow{}
    \mathbf{x} \sim p(\mathbf{x}| \mathbf{z}, \mathbf{y})
    $.
    \item \textbf{Reconstruction:} Similarly to standard VAE, we can reconstruct $\mathbf{x}$: 
    $ 
    \mathbf{y} \sim q(\mathbf{y}| \mathbf{x})
    \xrightarrow{}
    \mathbf{z} \sim q(\mathbf{z}| \mathbf{x})
    \xrightarrow{}
    \mathbf{x} \sim p(\mathbf{x}| \mathbf{z}, \mathbf{y})
    $.
    \item \textbf{Generative Reconstruction:} Additionally, we can reconstruct $\mathbf{x}$ by combining the generation and the reconstruction:
    $ 
    \mathbf{y}^{*} \sim q(\mathbf{y}^{*}| \mathbf{x})
    \xrightarrow{}
    \mathbf{u} \sim q(\mathbf{u}| \mathbf{y}^{*})
    \xrightarrow{}
    \mathbf{y} \sim p(\mathbf{y}| \mathbf{u})
    \xrightarrow{}
    \mathbf{z} \sim p(\mathbf{z}| \mathbf{y}, \mathbf{u}),
    \xrightarrow{}
    \mathbf{x} \sim p(\mathbf{x}| \mathbf{z}, \mathbf{y})
    $.
\end{itemize}

In order to highlight the super-resolution part in our model, we refer to it as the \textit{super-resolution VAE} (srVAE).
    
    \section{Experiments}
    \label{sec:experiments}
\begin{figure*}[t]
\centering
\scalebox{1.0}{
\begin{subfigure}{1.\textwidth}
    \includegraphics[width=1\linewidth]{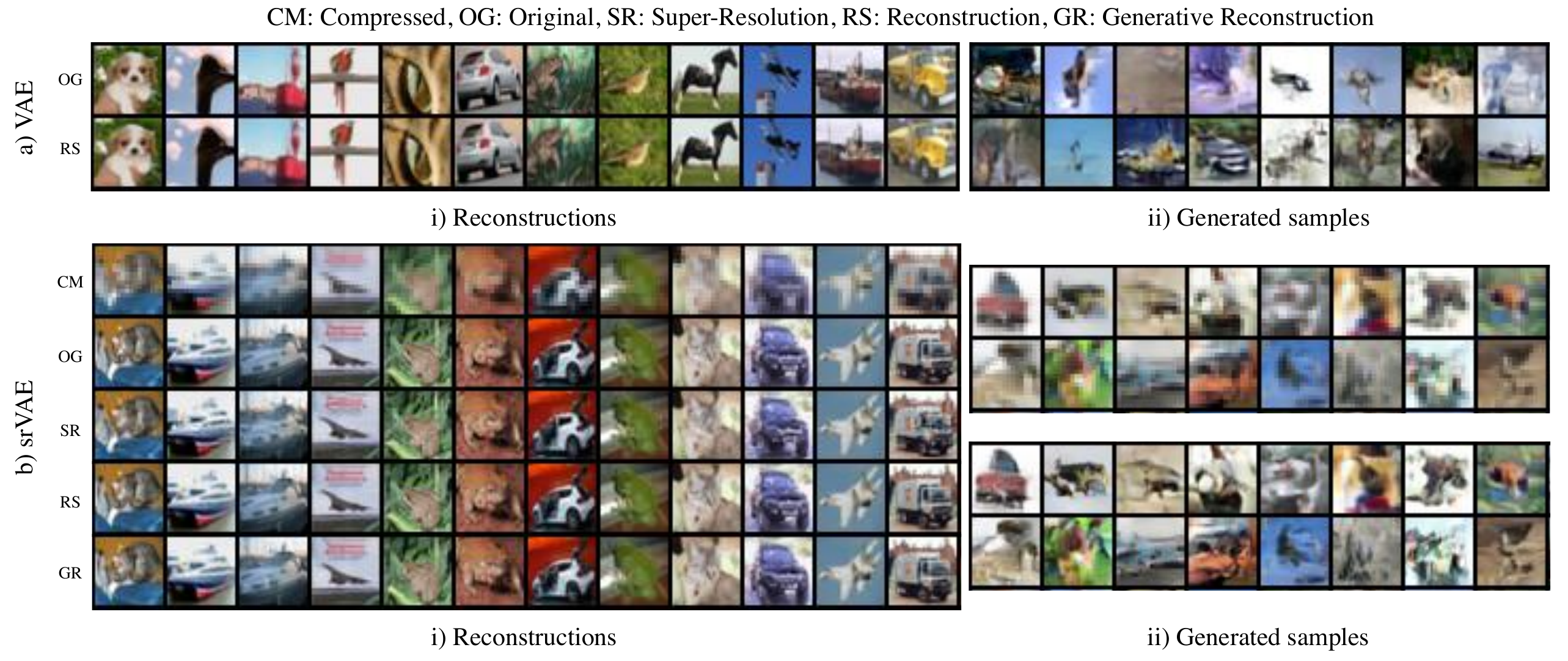}
\end{subfigure}}
\caption[CIFAR10]{Qualitative results of the VAE (a) and the srVAE (b) on CIFAR-10.}
\label{fig:cifar10}
\end{figure*}


\subsection{Setup}
We evaluated the following two models for density estimation; (i) a VAE (ii) the proposed two-level VAE (srVAE). 
In both models we employed RealNVP as a bijective prior \citep{dinh2016density}. 
Specifically, for our model, even though the compressed image $\mathbf{y}$ can be given by any deterministic and discrete transformation of the input data (i.e., the image label, grey-scale transformation, Fourier transform, sketch representation), we provide results with a $2\times$ downscaled image. 
The downscaled images still preserve the global structure of the samples while they disregard the high resolution details. 
Moreover, it will allow us to evaluate the model for its ability to perform super-resolution tasks. 
We set $K=M=16\times8\times8$ as latent dimentions in our experiments. 
The building blocks of the neural network implementation details are described in the Appendix, see Figure \ref{fig:nn_architecture}. 
We used a composition of DenseNets \citep{huang2016densely} and channel attention \citep{zhang2018image} with ELUs \citep{clevert2015fast} as activation functions.

We applied the proposed model to CIFAR-10 for quantitative and qualitative evaluation of natural images.
Additionally, we applied the model trained on CIFAR-10 to ImageNet32, without any additional fine-tuning in order to illustrate its adaption performance to a similar dataset. 
We evaluate the density estimation performance by \textit{bits per dimension} (bits/dim), $\mathcal{L}/(hwc \, \cdot \, \log(2))$, where $h$, $w$ and $c$ denote the height, width, and channels, respectively and we use the \textit{Fréchet Inception Distance} (FID) \citep{heusel2017gans} as a metric for image generation quality. 
The negative log-likelihood value (\textit{nll}) was estimated using $500$ weighted samples \citep{burda2015importance}, and for computing the FID scored we used $10$k generated images and $10$k real images from the test set, but also $50$k generated images and $50$k real images from the train set.

The code for this paper is available at \url{https://github.com/ioangatop/srVAE}.



\subsection{Evaluation on Natural Images}

\paragraph{Quantitative results} 
The density estimation and the image generation scores on CIFAR-10 and ImageNet32 are presented in Table \ref{tab:table_results}. 
Even though the VAE with the RealNVP prior follows an architecture without the use of any auto-regressive components and a single stochastic latent variable, it achieves a very competitive log-likelihood score (see Table \ref{tab:likelihood} in the Appendix). 
The importance of the data-driven prior is further supported by the low FID, as it manages to outperform various flow-based generative models (see the Appendix, Table \ref{tab:fid}). 
Finally, analyzing results in Table \ref{tab:table_results}, we make two observations. First, the $\mathrm{RE}_{x}$ is better in case of our approach. 
This result is to be expected since our model contains a super-resolution part. 
However, we pay a price for that, namely, we have an extra error coming from $\mathrm{RE}_{y}$. Second, both $\mathrm{KL}_{z}$ and $\mathrm{KL}_{u}$ are relatively large, and, thus, we claim the model does not suffer from the \textit{posterior collapse}. 
Interestingly, the $\mathrm{KL}$ part of the VAE is $1966$ and $1707$ for CIFAR-10 and ImageNet32, respectively, while our model achieves the sum of $\mathrm{KL}_{z}$ and $\mathrm{KL}_{u}$ around $1500$ on both datasets. 
This result suggests that introducing an additional random variable $\mathbf{y}$ helps to match the variational posteriors and the (conditional) priors, and the model does not bypass the latent variable $\textbf{z}$, verifying its importance.

Even though our model outperforms the VAE in the reconstruction loss of the original image ($\text{RE}_{x}$), due to the summation with the value of $\text{RE}_{y}$ it results in a poorer likelihood. 
However, we see that the srVAE significantly improves the FID score, as it produces more coherent and visually pleasing generations.



\paragraph{Qualitative results} 
We test the performance of the two models on image generation, reconstruction, and in the case of our model, additionally for conditional generation (super-resolution) and generative reconstruction tasks on CIFAR-10. The results are illustrated in Figure \ref{fig:cifar10}. 
The VAE with the bijective prior showcases an excellent performance on the natural image reconstruction task, which is contrary to the performance that is often provided in the literature. 
This maybe be associated with the effectiveness of a powerful, invertible, data-driven prior like RealNVP and its ability to boost the performance significantly with negligible sacrifice on generation speed, and none on inference. The provided unconditional generations, instead of being characterised as blurry, manage to output images with a coherent global structure. 

In contrast with the VAE, our approach breaks the generation of an image into a two-step process. 
It first generates a compressed sample through the latent variable $\mathbf{u}$, and then adds local structure with the help of the stochastic variable $\mathbf{z}$. 
The provided results illustrate that indeed the generations of the first step outputs an outline as a general concept which is then enriched with additional components, resulting in a sharp image. 
While a proportion of the generations of VAE tend to be noisy and abstract, the two-staged approach seems to generate smoother, higher fidelity results.
Moreover, due to our choice of the model, i.e, the compressed image as a $2\times$ downscaled representation, the unconditional generation functionality is essentially a super-resolution task. 
The model manages to perform accurate reconstructions of the original images, providing a novel generative approach.

More detailed results and analysis of the conducted experiments are provided in the Appendix \ref{sec:appendix_A3}.

    \section{Conclusion}
    \label{sec:section_5}
    We propose a new type of generative model which is able to perform both conditional and unconditional sampling which demonstrate improved quantitative performance in terms of FID of the generating sample on standard image modelling benchmark. 
In addition, we demonstrate that VAEs employed with a RealNVP prior can result in a competitive density estimation performance, despite its non-autoregressive architecture form and a single stochastic latent variable.
Our approach opens new directions in the VAE framework.
First, it allows usage of the log-likelihood-based objective to generate crisp images.
Second, the introduction of a downscaled image in the framework alleviates common issues in learning latent variables.
Third, it introduces the super-resolution into the VAE framework.
All these aspects could be further studied and developed to obtain better quality of generated images.

    \bibliography{main}
    \bibliographystyle{icml2020}

    \newpage
    \onecolumn
    \begin{asection}
    \section*{Appendix}
    \addcontentsline{toc}{section}{Appendix}
    \label{sec:appendix}
\subsection{Derivation of the lower bound}
\label{sec:appendix_A1}

Expanding the lower bound from (\ref{eq:elbo1}), from the first part we will have

\begin{gather*}
    \E\mathord{_{q(\mathbf{w})}}
    \Big[
    \log p(\mathbf{x}, \mathbf{w})
     \Big]
    =
    \E\mathord{_{q(\mathbf{w})}}
    \Big[
    \log 
    p(\mathbf{x}|\mathbf{y}, \mathbf{z}) \,
    p(\mathbf{z}|\mathbf{y}, \mathbf{u}) \,
    p(\mathbf{y}|\mathbf{u}) \,
    p(\mathbf{u})
    \Big]
    =
    \E\mathord{_{q(\mathbf{z}|\mathbf{y}, \mathbf{x}) q(\mathbf{y}|\mathbf{x})}}
    \Big[ \log p(\mathbf{x}|\mathbf{y}, \mathbf{z}) \Big]
    +\\+
    \E\mathord{_{q(\mathbf{z}|\mathbf{y}, \mathbf{x}) \,
    q(\mathbf{u}|\mathbf{y}) \, q(\mathbf{y}|\mathbf{x})}}
    \Big[ 
    \log
    p(\mathbf{z}|\mathbf{y}, \mathbf{u})
    \Big]
    +
    \E\mathord{_{q(\mathbf{u}|\mathbf{y}) q(\mathbf{y}|\mathbf{x})}}
    \Big[ \log p(\mathbf{y}|\mathbf{u}) \Big] \,
    + 
    \E\mathord{_{q(\mathbf{u}|\mathbf{y}) q(\mathbf{y}|\mathbf{x})}}
    \Big[ \log p(\mathbf{u}) \Big],
\end{gather*}

and for the second

\begin{align*}
    \E\mathord{_{q(\mathbf{w})}}
    \Big[
    \log q(\mathbf{w})
    \Big]
    =
    \E\mathord{_{q(\mathbf{z}|\mathbf{y}, \mathbf{x})}}
    \Big[ 
    \log
    q(\mathbf{z}|\mathbf{y}, \mathbf{x})
    \Big]
    +
    \E\mathord{_{q(\mathbf{y}|\mathbf{x})}}
    \Big[ \log q(\mathbf{y}|\mathbf{x}) \Big] \,
    +
    \E\mathord{_{q(\mathbf{u}|\mathbf{y}) \, q(\mathbf{y}|\mathbf{x})}}
    \Big[ \log q(\mathbf{u}|\mathbf{y}) \Big].
\end{align*}

Interestingly, plugging the above terms back to (\ref{eq:elbo1}) and rearranging them, we will have

\begin{gather*}
    \mathcal{L}(\mathbf{x})
    \overset{(\ref{eq:elbo1})}{=}
    \underbrace{
    \E\mathord{_{q(\mathbf{z}|\mathbf{y}, \mathbf{x}) q(\mathbf{y}|\mathbf{x})}}
    \Big[ \log p(\mathbf{x}|\mathbf{y}, \mathbf{z}) \Big] \,
    +
    \E\mathord{_{q(\mathbf{z}|\mathbf{y}, \mathbf{x}) \,
    q(\mathbf{u}|\mathbf{y}) \, q(\mathbf{y}|\mathbf{x})}}
    \Big[ 
    \log
    p(\mathbf{z}|\mathbf{y}, \mathbf{u})
    \Big] - \E\mathord{_{q(\mathbf{z}|\mathbf{y}, \mathbf{x})}}
    \Big[ 
    \log
    q(\mathbf{z}|\mathbf{y}, \mathbf{x})
    \Big]
    }_{\boldsymbol{A}}
    +
    \\+
    \underbrace{
    \E\mathord{_{q(\mathbf{u}|\mathbf{y}) q(\mathbf{y}|\mathbf{x})}}
    \Big[ \log p(\mathbf{y}|\mathbf{u}) \Big] \,
    + 
    \E\mathord{_{q(\mathbf{u}|\mathbf{y}) q(\mathbf{y}|\mathbf{x})}}
    \Big[ \log p(\mathbf{u}) \Big] \,
    -
    \E\mathord{_{q(\mathbf{y}|\mathbf{x})}}
    \Big[ \log q(\mathbf{y}|\mathbf{x}) \Big] \,
    -
    \E\mathord{_{q(\mathbf{u}|\mathbf{y}) q(\mathbf{y}|\mathbf{x})}}
    \Big[ \log q(\mathbf{u}|\mathbf{y}) \Big]
    }_{\boldsymbol{B}}.
\end{gather*}

Working with term $\boldsymbol{B}$, one can see that

\begin{gather*}
    \boldsymbol{B}
    = 
    \E\mathord{_{q(\mathbf{u}|\mathbf{y}) q(\mathbf{y}|\mathbf{x})}}
    \Big[
    \log 
    \frac{p(\mathbf{y}|\mathbf{u}) p(\mathbf{u})}{q(\mathbf{u}|\mathbf{y}) q(\mathbf{y}|\mathbf{x})}
    \Big],
\end{gather*}

which denotes a (hidden) lower bound on of the marginal $\log p(\mathbf{y})$
with variational posterior $q(\mathbf{u}|\mathbf{y}) q(\mathbf{y}|\mathbf{x})$.

Thus, the resulted lower bound of the marginal likelihood of $\mathbf{x}$ would be

\begin{align*}
    \mathcal{L}(\mathbf{x})
    &=
    \E\mathord{_{q(\mathbf{z}| \mathbf{x}, \mathbf{y}) \, q(\mathbf{y}|\mathbf{x})}} \log 
    p_{\theta}(\mathbf{x}| \mathbf{y}, \mathbf{z}) \,  
    - \KL({q(\mathbf{z}| \mathbf{x}, \mathbf{y}) || p(\mathbf{z}| \mathbf{y}, \mathbf{u}}))
    + \E\mathord{_{q(\mathbf{u})}} \log 
    p_{\theta}(\mathbf{y}| \mathbf{u})
    -
    \KL({q(\mathbf{u}| \mathbf{y}})||p(\mathbf{u}) )
    \\&=
    \E\mathord{_{q(\mathbf{z}| \mathbf{x}, \mathbf{y}) \, q(\mathbf{y}|\mathbf{x})}} \log 
    p_{\theta}(\mathbf{x}| \mathbf{y}, \mathbf{z}) \,  
    - \KL({q(\mathbf{z}| \mathbf{x}, \mathbf{y}) || p(\mathbf{z}| \mathbf{y}, \mathbf{u}}))
    +
    \E\mathord{_{q(\mathbf{u}|\mathbf{y}) q(\mathbf{y}|\mathbf{x})}}
    \Big[
    \log 
    \frac{p(\mathbf{y}|\mathbf{u}) p(\mathbf{u})}{q(\mathbf{u}|\mathbf{y}) q(\mathbf{y}|\mathbf{x})}
    \Big].
\end{align*}

\subsection{Neural Network Architecture}
\label{sec:appendix_A2}

In Figure \ref{fig:nn_architecture} is depicted the main architecture of the VAE with the bijective prior as well as the optimization choices. The design choices for the encoder and decoder form the building blocks to every model that was trained and evaluated.

\begin{figure}[htbp]
    \centering
    \scalebox{.4}{
    \begin{subfigure}{1.\textwidth}
      \centering
      \includegraphics[width=1.\linewidth]{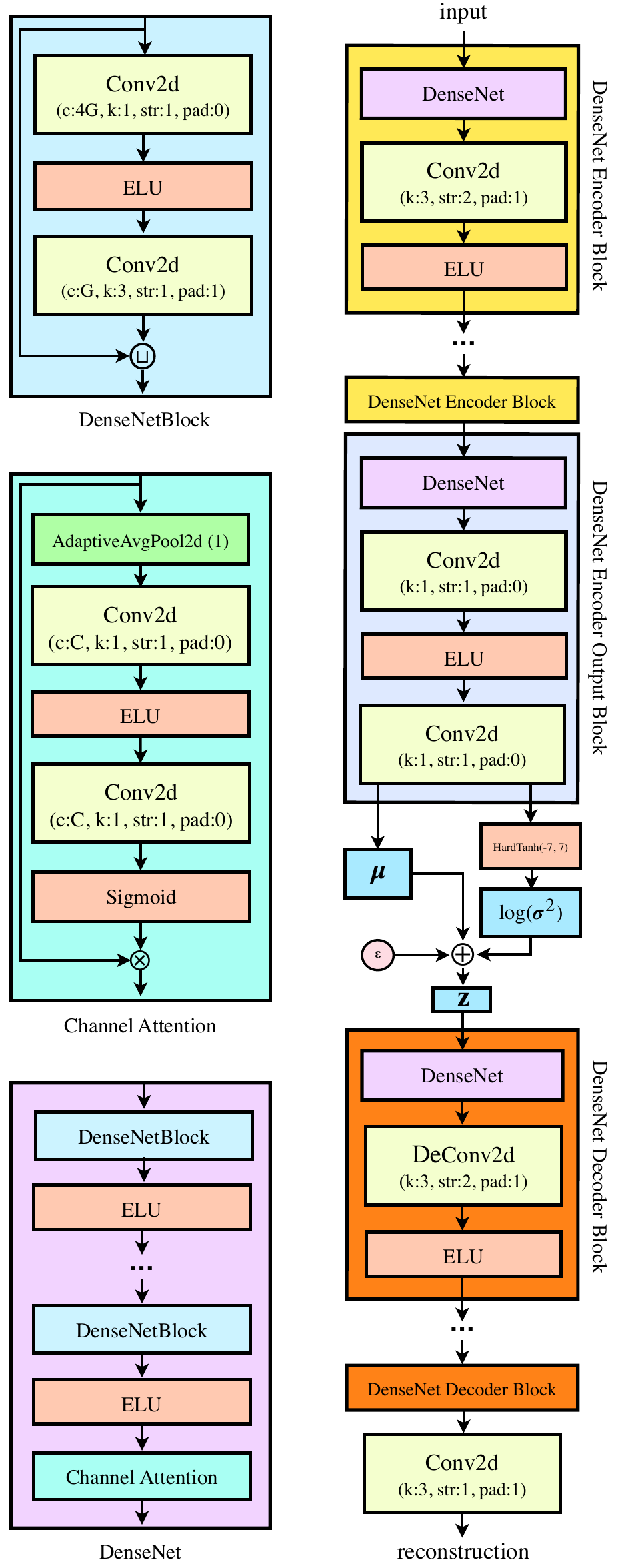}
    \end{subfigure}}
    \caption[Architecture of the autoencoder]{Architecture of our autoencoder. On the right, there are some basic buildings block of the network. The notation as 'G' on the Conv2D channels indicate the growth rate of the densely connected network. The $\epsilon$ indicates a random variable drawn from a standard Gaussian, which helps us to make use of the \textit{reparametrization} trick. Until $\mathbf{z}$, we refer to this architecture as \textit{Encoder NN} and thereafter as \textit{Decoder NN}. The former and the later form the \textit{building blocks} to every model that we train and evaluate. All models that are evaluated had $\sim 35\mathrm{M}$ trainable parameters, were trained for $2$ thousands epochs, using AdaMax optimizer \citep{kingma2014adam} and the dimensionality of all the latent variables kept at $8 \times 8 \times 16$. We applied weight normalization on the parameters with data-depended initialisation \citep{salimans2016weight}.}
    \label{fig:nn_architecture}
\end{figure}

\subsection{Supplementary results}
\label{sec:appendix_A3}
The datasets CIFAR-10 and ImageNet32 were split as described in \cite{hoogeboom2019integer}. We notice that some papers in the literature use different and, in our opinion, unfair data division.


\paragraph{Quantitative Results}
\label{sec:appendix_A31}

Additional quantitative results are presented in Tables \ref{tab:likelihood} and \ref{tab:fid}.


\begin{table}[t]
\caption{Generative modelling performance in bits per dimension. The symbol $^{\dag}$ on the ImageNet32 data denotes that the obtained results were produced using a different downsampling method from the one that was introduced by \cite{oord2016pixel}, indicating  not a fair comparison with the other methods. In the case of Flow++, we provide the results of the variational dequantization in the brackets.}
\begin{center}
\scalebox{1.0}{
\begin{tabular}{clcc}
    \textbf{Model Family} & \multicolumn{1}{c}{\textbf{Model}} & \textbf{CIFAR-10} & \textbf{ImageNet 32x32} \\ \hline
    \multirow{7}{*}{Autoregressive}
     & PixelCNN \citep{oord2016pixel} & 3.14 & –\\
     & PixelRNN \citep{oord2016pixel} & 3.00 & 3.86 \\
     & Gated PixelCNN \citep{oord2016conditional} & 3.03 & 3.83 \\
     & PixelCNN++ \citep{salimans2017pixelcnn} & 2.92 & – \\
     & Image Transformer \citep{parmar2018image} & 2.90 & 3.77 \\
     & PixelSNAIL \citep{chen2017pixelsnail} & 2.85 & 3.80  \\ \hline
    \multirow{8}{*}{Non-autoregressive} & RealNVP \citep{dinh2016density} & 3.49 & 4.28  \\
     & DVAE++ \citep{vahdat2018dvae} & 3.38 & –  \\
     & Glow \citep{kingma2018glow} & 3.35 & 4.09  \\
     & IAF-VAE \citep{kingma2016iaf} & 3.11 & –  \\
     & BIVA \citep{maale2019biva} & 3.08 & 3.96 \\
     & Flow++ \citep{ho2019flow} & \hspace{24pt} 3.29 (3.08)  & \hspace{24pt}  -- (3.86)  \\
     & VAE with bijective prior (ours) &  3.51 & 3.80$^{\dag}$  \\
     & srVAE (ours) &  3.65 & 4.00$^{\dag}$ \\ \bottomrule
    \end{tabular}}
\label{tab:likelihood}
\end{center}
\end{table}


\begin{table*}[ht]
\caption{FID scores obtained from different models trained on CIFAR-10. Lower FID implies better sample quality. All results except ours are taken from \cite{chen2019residual}. In the case of our VAEs, we provide the values obtained on the test set and the training set (in brackets).}
\centering
\scalebox{1.0}{
  \begin{tabular}{lc}
    \textbf{Model} & \textbf{FID} \\
    \toprule
    PixelCNN \citep{oord2016conditional} & 65.93 \\
    PixelIQN \citep{ostrovski2018autoregressive} & 49.46 \\
    \hline
    iResNet Flow \citep{liang2017learning} & 65.01 \\
    GLOW \citep{kingma2018glow} & 46.90   \\
    Residual Flow \citep{chen2019residual} & 46.37 \\
    \hline
    DCGAN \citep{radford2015unsupervised} & 37.11   \\
    WGAN-GP \citep{gulrajani2017improved} & 36.40 \\
    \hline
    VAE with bijective prior (ours) & 41.36 (37.25) \\
    srVAE (ours) & 34.71 (29.95) \\
    \bottomrule
  \end{tabular}}
  \label{tab:fid}
\end{table*}


\paragraph{Qualitative Results}
\label{sec:appendix_A32}

Additional qualitative results of the VAE and the srVAE for CIFAR-10 are illustrated in Figures \ref{fig:vae_cifar10}, \ref{fig:two_staged_1}, \ref{fig:two_staged_2} and \ref{fig:comp_cifar10}. For ImageNet32, the images are presented in Figures \ref{fig:vae_recon_imagenet}, \ref{fig:super_two_staged_imagenet} and \ref{fig:comp_imagenet}.

\begin{figure}[ht]
    \centering
    \scalebox{.98}{
    \begin{subfigure}{1.\textwidth}
      \centering
      \includegraphics[width=1.\linewidth]{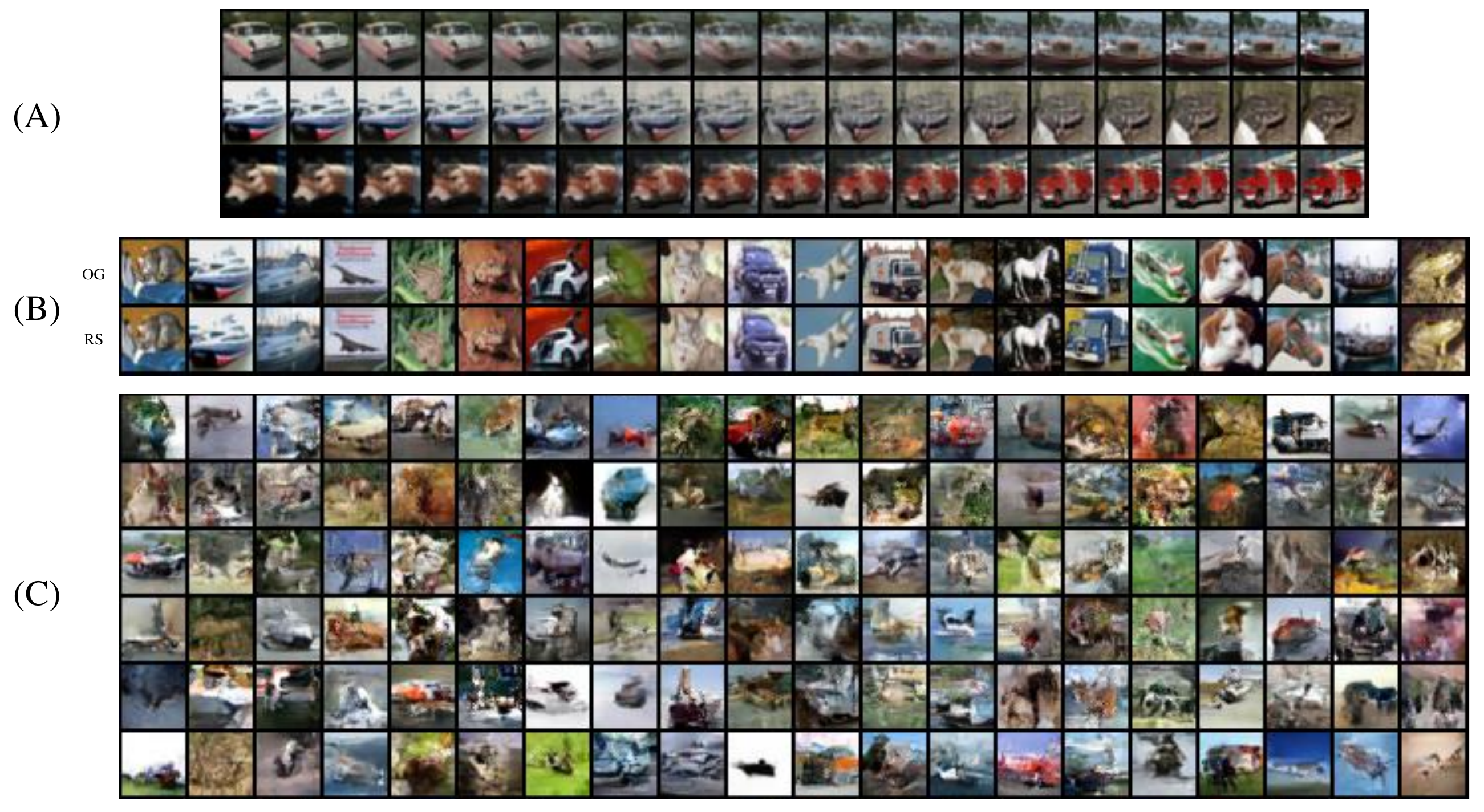}
    \end{subfigure}}
    \caption{Qualitative results on CIFAR-10 of the VAE with the bijective prior. (A) Interpolation (B) Reconstructions (OG: real images, RS: reconstructions) and (C) Unconditional Generations.}
    \label{fig:vae_cifar10}
\end{figure}

\begin{figure}[ht]
    \centering
    \scalebox{0.95}{
    \begin{subfigure}{1.\textwidth}
      \centering
      \includegraphics[width=1.\linewidth]{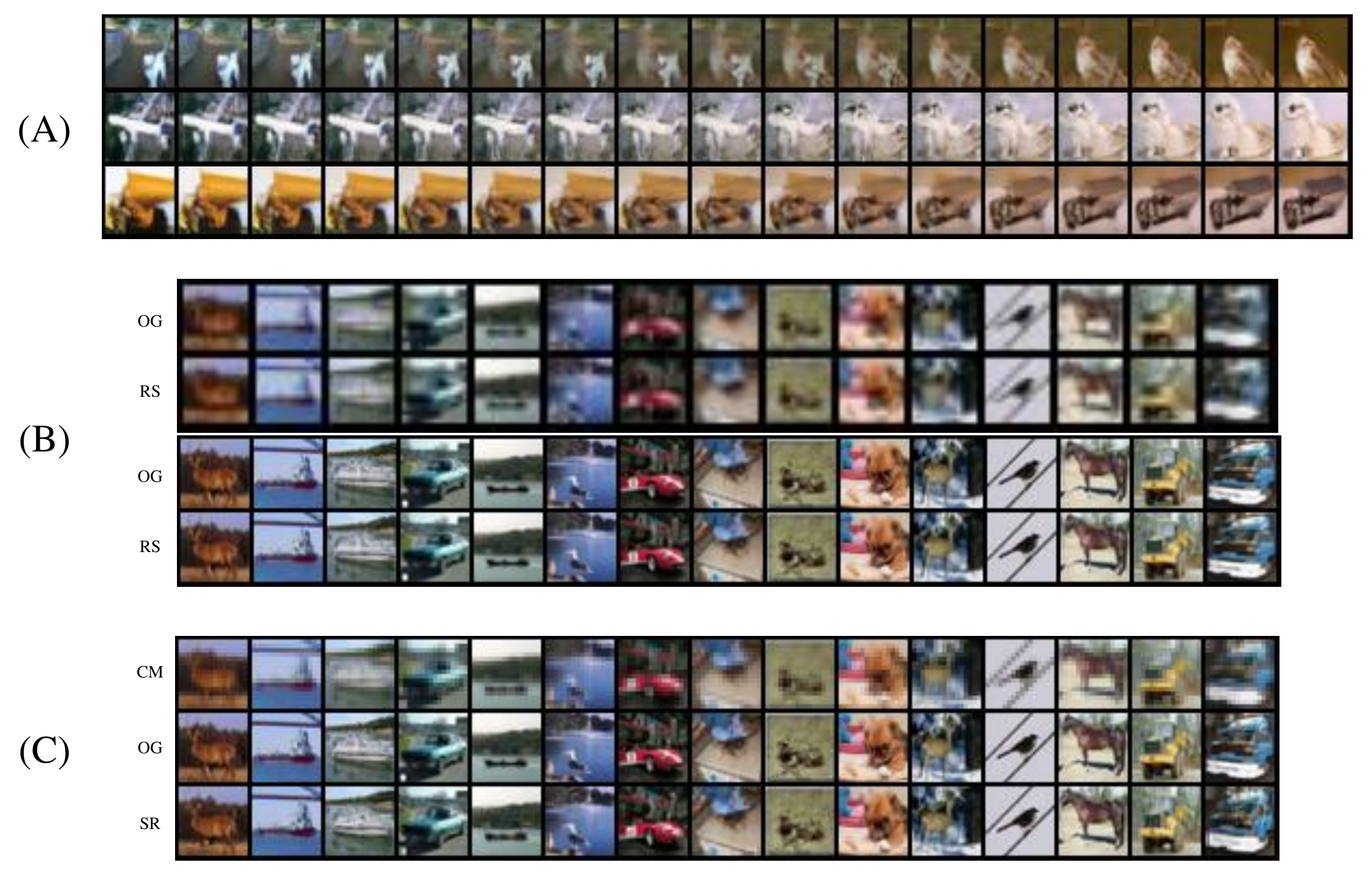}
    \end{subfigure}}
    \caption{Qualitative results on CIFAR-10 of the srVAE. (A) Interpolation (B) Reconstructions (OG: real images, RS: reconstructions) and (C) Super-Resolution (CM: downscaled images, OG: real images, RS: conditional generations (super-resolution).}
    \label{fig:two_staged_1}
\end{figure}

\begin{figure}[ht]
    \centering
    \scalebox{.8}{
    \begin{subfigure}{1.\textwidth}
      \centering
      \includegraphics[width=1.\linewidth]{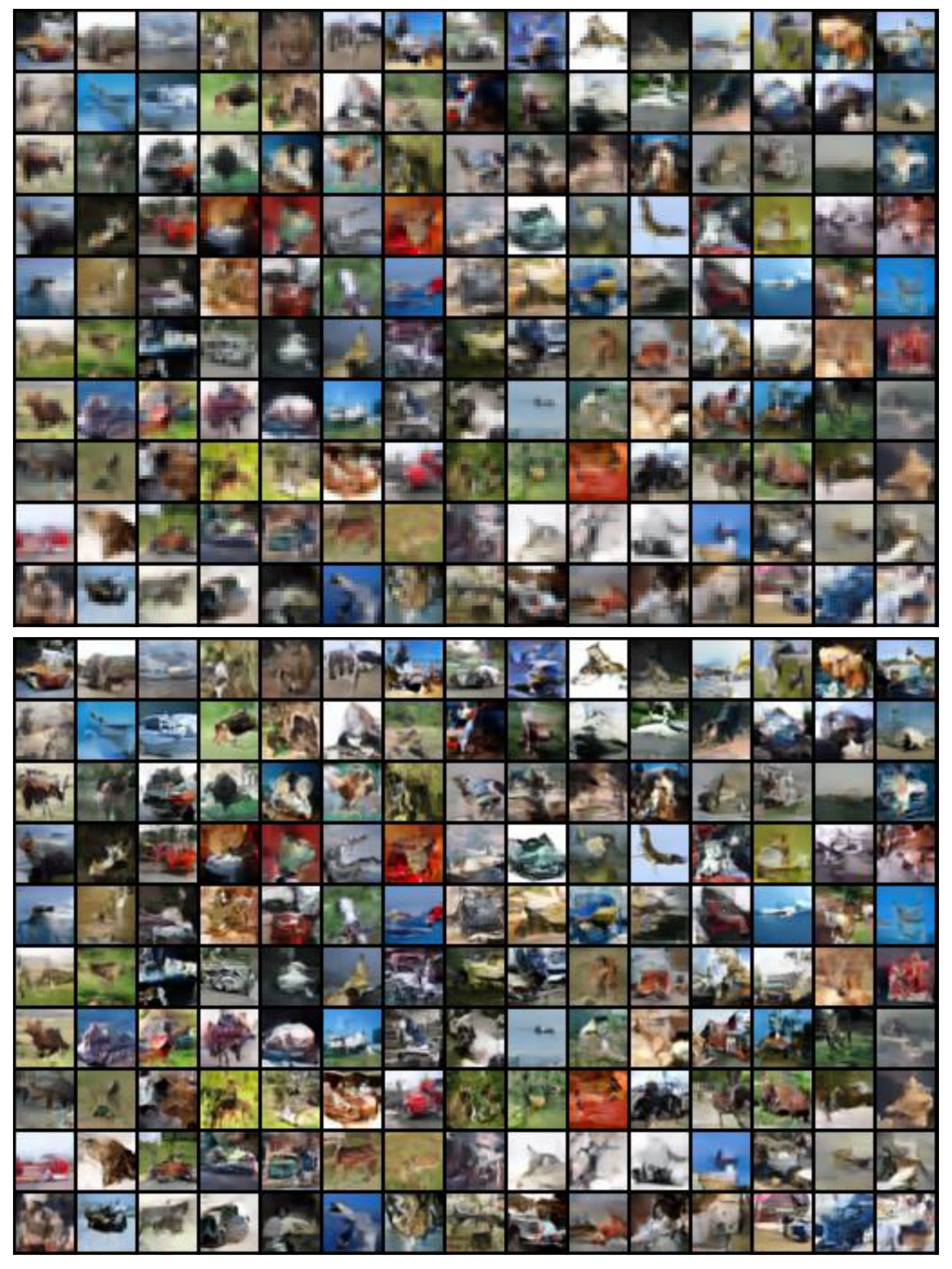}
    \end{subfigure}}
    \caption[]{Unconditional generations of the srVAE trained on CIFAR10. The model initially generates the $\times2$ downscaled (compressed) representation of the image which intends to captures the "global" structure and then adds the local structure (the "details") while increasing its receptive field. These results indicate that the model successfully captures  the global structured information from data on the first stage with the latent variable $\mathbf{u}$ and then adds a local structure with the help of the latent variable $\mathbf{z}$, whose responsibility is to capture the missing information between the compressed and the original data.}
    \label{fig:two_staged_2}
\end{figure}

\begin{figure}[ht]
    \centering
    \scalebox{.8}{
    \begin{subfigure}{1.\textwidth}
      \centering
      \includegraphics[width=1.\linewidth]{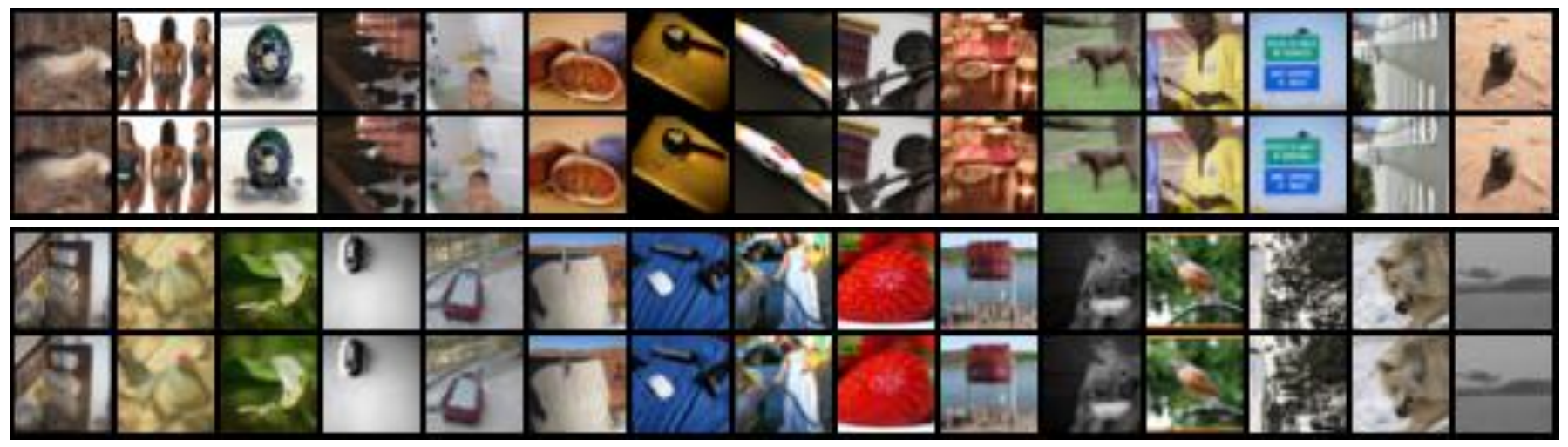}
    \end{subfigure}}
    \caption[]{Reconstruction results on $\text{ImageNet}^{\dag}$ from the VAE with the bijective prior.  The results show that the model can successfully reconstruct natural images from a different source though its $8\times8\times16$ dimensional latent space. Top row indicates the ground truth samples and the second rows present the results after reconstruction.}
    \label{fig:vae_recon_imagenet}
\end{figure}

\begin{figure}[ht]
    \centering
    \scalebox{.8}{
    \begin{subfigure}{1.\textwidth}
      \centering
      \includegraphics[width=1.\linewidth]{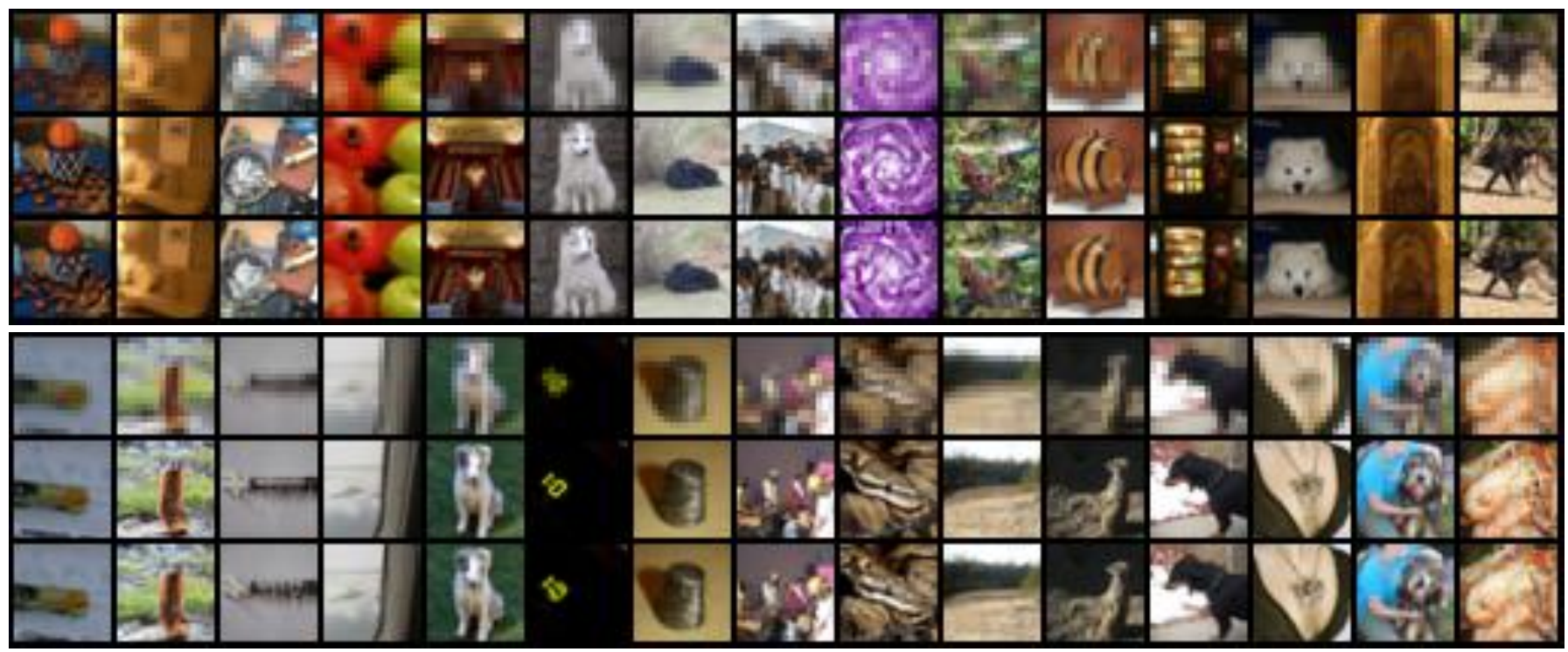}
    \end{subfigure}}
    \caption[]{Super-Resolution results of the srVAE on $\text{ImageNet}^{\dag}$. Even though the model was trained on CIFAR10, its performance showcases its robustness capabilities. The top, second and third row illustrate the $2\times$ downscaled image, the original and the after the super resolution result of the model, respectively.}
    \label{fig:super_two_staged_imagenet}
\end{figure}

\begin{figure}[ht]
    \centering
    \scalebox{.8}{
    \begin{subfigure}{1.\textwidth}
      \centering
      \includegraphics[width=1.\linewidth]{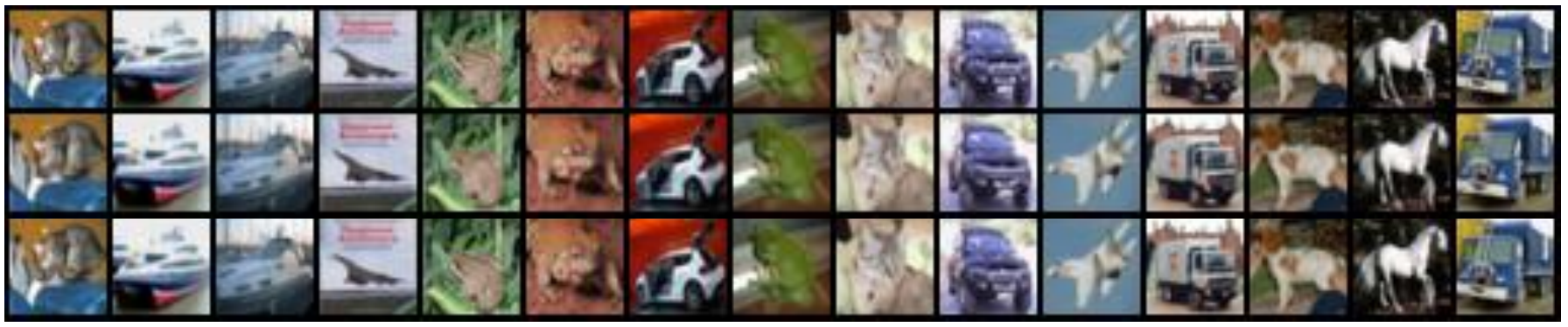}
    \end{subfigure}}
    \caption[]{Comparison on the image reconstruction on CIFAR10 between the VAE with the bijective prior (second row) and the srVAE (third row). Even though both models output images very similar to the original one (top row), the srVAE seems to preserve more details from the ground truth (take a look at images 5, 10 and 15).}
    \label{fig:comp_cifar10}
\end{figure}

\begin{figure}[ht]
    \centering
    \scalebox{.8}{
    \begin{subfigure}{1.\textwidth}
      \centering
      \includegraphics[width=1.\linewidth]{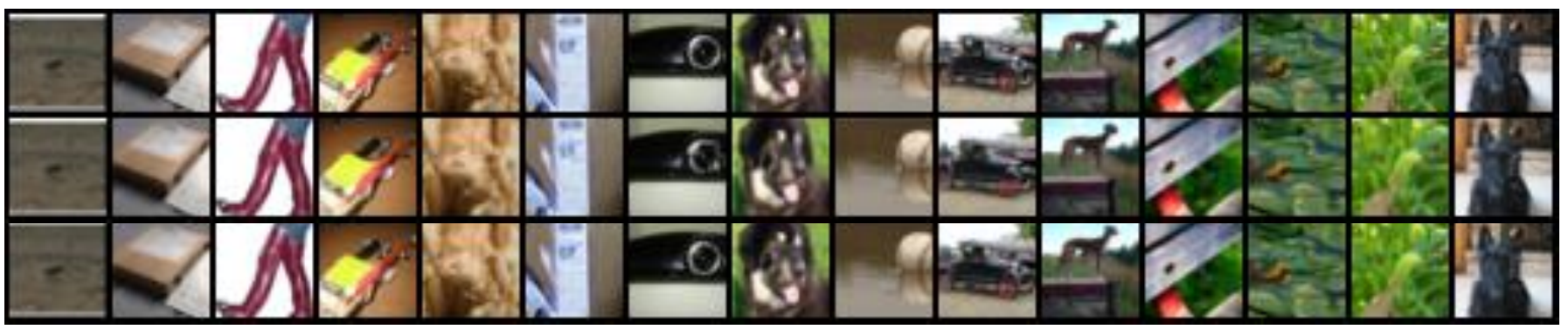}
    \end{subfigure}}
    \caption[]{Comparison on the image reconstruction on $\text{ImageNet}^{\dag}$ between the VAE with the bijective prior (second row) and the srVAE (third row). Again, we can notice that the last row seems to be more similar to the ground truth samples (top row).}
    \label{fig:comp_imagenet}
\end{figure}

    \end{asection}

\end{document}